\documentclass[11pt,a4paper]{article}
\usepackage[hyperref]{emnlp2020}
\usepackage{times}
\usepackage{latexsym}

\usepackage{tipa}

\usepackage{bm}

\usepackage{caption}
\usepackage{subcaption}
\usepackage{comment}
\usepackage{microtype}
\usepackage{amsmath}
\usepackage{amsfonts}
\usepackage{booktabs}
\usepackage{amsthm}
\usepackage{graphicx}
\usepackage{bbm}
\usepackage{multirow}
\usepackage{commath}
\usepackage{csquotes}
\usepackage{cancel}
\usepackage{xcolor}
\usepackage{inconsolata}

\usepackage{graphicx} 
\graphicspath{{./img/}}

\usepackage{booktabs}
\usepackage{cleveref}
\crefname{section}{\S}{\S\S}
\Crefname{section}{\S}{\S\S}
\crefname{table}{Table}{Tables}
\crefname{figure}{Figure}{Figures}
\crefname{algorithm}{Algorithm}{}
\crefname{equation}{eq.}{}
\crefname{appendix}{App.}{}
\crefname{prop}{Proposition}{}
\crefformat{section}{\S#2#1#3}  %

\usepackage{todonotes}

\newcommand{\citeposs}[1]{\citeauthor{#1}'s (\citeyear{#1})}

\newcommand*\iftodonotes{\if@todonotes@disabled\expandafter\@secondoftwo\else\expandafter\@firstoftwo\fi}  %
\makeatother

\usepackage{colortbl}

\newcommand{\vs}{\mathbf{s}}
\newcommand{\vc}{\mathbf{c}}
\newcommand{\vp}{\mathbf{p}}

\newcommand{\vh}{\mathbf{h}}

\newcommand{\vmu}{{\boldsymbol \mu}}

\newcommand{\BERT}{\mathrm{BERT}}
\newcommand{\mask}{\textsc{mask}}
\newcommand{\normal}{\mathcal{N}}
\newcommand{\ent}{\mathrm{H}}
\newcommand{\MI}{\mathrm{I}}
\newcommand{\calM}{\mathcal{M}}
\newcommand{\calC}{\mathcal{C}}
\newcommand{\calW}{\mathcal{W}}
\newcommand{\eos}{\textsc{eos}}
\newcommand{\bos}{\textsc{bos}}

\newcommand{\defn}[1]{\textbf{#1}}
\newcommand{\unigram}{p_{\textit{unigram}}}
\newcommand{\word}[1]{\textit{#1}}
\newcommand{\wordnet}{\textit{WordNet}}
\newcommand{\softmax}{\mathrm{softmax}}
\newcommand{\diag}{\mathrm{diag}}

\aclfinalcopy %
\def\aclpaperid{644} %

\title{Speakers Fill Lexical Semantic Gaps with Context}%

\newcommand{\ucambridge}{\normalfont \text{\textipa{D}}}
\newcommand{\ethz}{\text{\normalfont \textipa{Q}}}
\newcommand{\uharvard}{\normalfont \text{\textipa{@}}}
\newcommand{\mpi}{\normalfont \text{\textipa{S}}}
\newcommand{\hse}{\normalfont \text{\textipa{K}}}

\author{Tiago Pimentel$^{\ucambridge}$~\;~ Rowan Hall Maudslay$^{\ucambridge}$~\;~ Dami\'an Blasi$^{\uharvard,\mpi,\hse}$~\;~ Ryan Cotterell$^{\ucambridge,\ethz}$ \\
  $^{\ucambridge}$University of Cambridge~\;~$^{\uharvard}$Harvard University~\;~$^{\mpi}$MPI SHH~\;~$^{\hse}$HSE University~\;~%
  $^{\ethz}$ETH Z\"{u}rich\\
  \texttt{\href{mailto:tp472@cam.ac.uk}{tp472@cam.ac.uk}}~\;~ \texttt{\href{mailto:rh635@cam.ac.uk}{rh635@cam.ac.uk}} \\
  \texttt{\href{mailto:dblasi@fas.harvard.edu}{dblasi@fas.harvard.edu}}~\;~ \texttt{\href{mailto:ryan.cotterell@inf.ethz.ch}{ryan.cotterell@inf.ethz.ch}}
}

\date{}

\begin{document}
\maketitle
\begin{abstract}
Lexical ambiguity is widespread in language, allowing for the reuse of economical word forms and therefore making language more efficient.
If ambiguous words cannot be disambiguated from context, however, this gain in efficiency might make language less clear---resulting in frequent miscommunication. 
For a language to be clear \emph{and} efficiently encoded, 
we posit that the lexical ambiguity of a word type should correlate with 
how much information context provides about it, on average.
To investigate whether this is the case, we operationalise the lexical ambiguity of a word as the entropy of meanings it can take, and provide two ways to estimate this---one which requires human annotation (using \wordnet{}), and one which does not (using $\BERT$),
making it readily applicable to a large number of languages.
We validate these measures by showing that, on six high-resource languages, there are significant 
Pearson correlations between our $\BERT$-based estimate of ambiguity and the number of synonyms a word has in \wordnet{} (e.g.\ $\rho = 0.40$ in English).
We then test our main hypothesis---that a word's lexical ambiguity should negatively correlate with its contextual uncertainty---and find significant correlations on all 18 typologically diverse languages we analyse. This suggests that, in the presence of ambiguity, speakers compensate by making contexts more informative.\looseness=-1 
\end{abstract}

\section{Introduction} \label{sec:introduction}

Linguistic structure and meaning are often underdetermined in the linguistic signal.
In an extreme case this can lead to \defn{ambiguity}:
sentences might allow more than one valid syntactic structure, 
and pronouns could corefer to various antecedents.
Complementarily, linguistic signals can also overdetermine some aspect of the intended message---for instance, agreement patterns may require redundant marking,
and word forms might occupy sparsely populated parts of the phonological space \cite{harley1998causes}.
In a tradition that goes back at least to \citeauthor{zipf1949human}, it has been hypothesised that 
individuals maintain an efficient balance between over- and under-specifying an intended message.
Such balance is mediated by conflicting pressures for both \defn{clarity} (the quality that allows the reconstruction of the intended message), and \defn{economy of expression} (which allows for inexpensive and rapid encoding of the message in a linguistic signal). 

\begin{figure}
    \centering
    \includegraphics[trim={0 .55cm 0 .45cm},clip,width=\columnwidth]{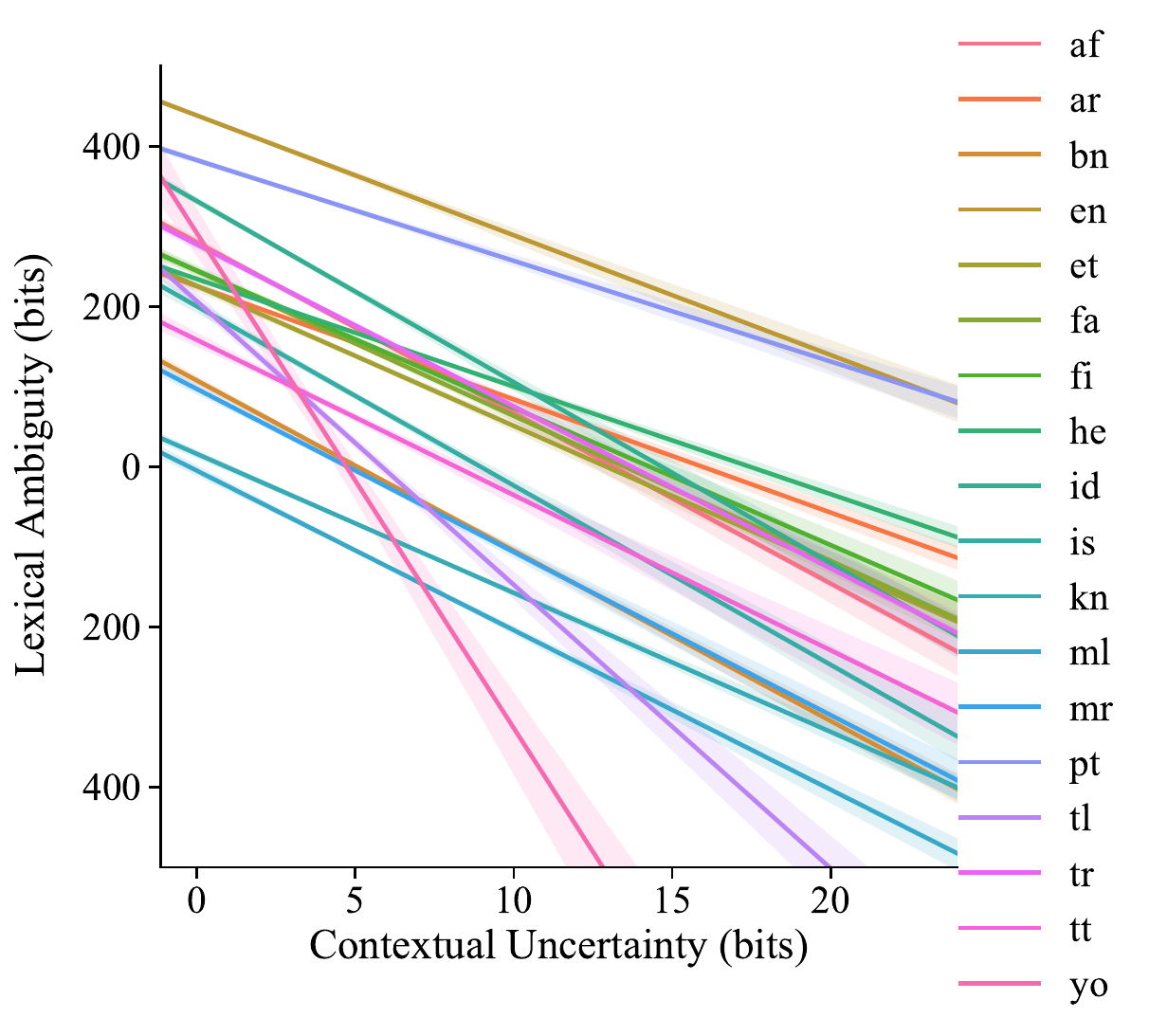}
    \caption{The relationship between contextual uncertainty---how uncertain a word is given its context---and lexical ambiguity, across a diverse set of languages.\looseness=-1}
    \label{fig:mi_full}
\end{figure}

A recent instantiation of this idea is that in an efficient language, one expects economical words (which are short or phonotactically simple) to be associated with multiple unrelated meanings, so they can be more widely used \citep{piantadosi2012communicative}. 
At first blush, this may appear to sacrifice clarity, increasing ambiguity and making it more difficult for a listener to resolve the linguistic signal.
The emerging picture from psycholinguistics and pragmatics, however, is that individuals \textit{can} fill in these ambiguous gaps, by tapping on additional linguistic or extra-linguistic cues \cite{tanenhaus1995integration,federmeier1999rose,dautriche2018learning}.
An obvious example is given by the role of contextual information in reducing the ambiguity associated with the meaning of a word form. 
For instance, the contexts which surround the word \word{ruler} in the sentences `Alice borrowed a \word{ruler} from her friends at school' and `Bob rose to power and became a ruthless \word{ruler}' each play a crucial role in disambiguating its intended underlying meaning.

To remain robust in the presence of noise, 
we may expect the linguistic signal to be on average somewhat overdetermined by the speaker, 
leading to redundancy in how words and their contexts determine the intended meaning.\footnote{We refer to overdetermination with relation to redundancies in the signal itself, rather than a precise intended meaning.}
By analysing this redundant information-theoretically under the assumption that languages strike a balance between economy of expression and clarity, we derive that the `amount' of lexical ambiguity in a given word type should negatively correlate with how uncertain on average the word is given its context (see \cref{sec:robustness}).
As communication unfolds, the efficiency of a particular word can only be modestly modified \cite[e.g.\ by choosing clipped forms when available;][]{mahowald2013info}. 
However, contexts can be enriched or demoted dynamically, so as to complement a word with the evidence needed for disambiguation.

To investigate whether it is the case that the contexts in which a word appears are systematically adapted to enable disambiguation, we first provide an operationalisation of lexical ambiguity, grounded in information theory.
We then provide two methods for estimating it, one  using \wordnet{} \cite{wordnet}, and the other using multilingual $\BERT{}$'s contextualised embeddings \citep{devlin2019bert}, which allows us to explore a large set of languages. 
We validate our lexical ambiguity measurements by comparing one to the other in six high-resource languages from four language families (Afro-Asiatic: Arabic; Austronesian: Indonesian; Indo-European: English, Persian and Portuguese; Uralic: Finnish), and find significant correlations between the number of synsets in \wordnet{} and our $\BERT$ estimate (e.g.\ $\rho = 0.40$ in English), indicating that our annotation-free method for measuring lexical ambiguity is useful.
We then test our main hypothesis---that the contextual uncertainty about a word should negatively correlate with its degree of lexical ambiguity. 
First, we test this on the same set of six high-resource languages for which we have \wordnet{} annotation, and find significant negative correlations on five of them.
We then extend our evaluation, using our $\BERT{}$-based measure, to cover a much more representative set of 18 typologically diverse languages: Afrikaans, Arabic, Bengali, English, Estonian, Finnish, Hebrew, Indonesian, Icelandic, Kannada, Malayalam, Marathi, Persian, Portuguese, Tagalog, Turkish, Tatar, and Yoruba.\footnote{We refer to these using ISO 639-1 codes.}
In this set, we find significant negative correlations for all languages (see \cref{fig:mi_full}).

\section{Ambiguity in Language}

While the pervasiveness of ambiguity in language encumbers the algorithmic processing of natural language \citep{church, manning}, people seamlessly overcome ambiguity through both linguistic and non-linguistic means.
World knowledge, pragmatic inferences, and expectations about discourse coherence all contribute to rapidly decoding the intended message out of potentially ambiguous signals \cite{wasow2015ambiguity}. 
While sometimes ambiguity might indeed result in an observed processing burden \cite{frazier}, which could lead communication astray, individuals can in response retrace and reanalyse their inferences (as it has been famously shown in garden-path sentences like ``The horse raced past the barn fell''; \citeauthor{bever1970cognitive}, \citeyear{bever1970cognitive}).

This outstanding capacity to navigate ambiguous linguistic signals calls for a reexamination of the presence of ambiguity found in language.
If the linguistic signal was deterministically and uniquely decodable---as, for instance, in the universal language proposed by \citeauthor{wilkins1668essay} \cite{borges1937analytical}---then all of the para-linguistic evidence would be redundant, and the code underlying the signal would be substantially more cumbersome. 
On the other hand, if linguistic signals present individuals with too many compatible inferences, communication would break down. 
An extreme case is represented by Louis Victor Leborgne, an aphasia patient described by Paul Broca \cite{mohammed2018louis}.
Louis, in spite of immaculate comprehension and mental functions, was unable to utter anything else than the syllable ``tan'' in his attempts to communicate.
\looseness=-1

The most influential explanation offered for why natural languages are seemingly far from both extremes derives from the seminal work of \newcite{zipf1949human}. In that work, \citeauthor{zipf1949human} proposed several aspects of human cognition and behaviour could be derived from the principle of least effort. 
Languages should aim to minimise the complexity and cost of linguistic signals as much as possible, under the sole constraint that the signal can be decoded efficiently.
\looseness=-1

\vspace{-5pt}
\subsection{Lexical Ambiguity}
\vspace{-2pt}

We are concerned exclusively with \defn{lexical ambiguity}.
A classic example is the English word \word{bank}, which can refer to either an establishment where money is kept, or the patch of land alongside a river.  %
A significant source of lexical ambiguity is word types which exhibit multiple senses, which are said to be \defn{polysemous}
or \defn{homonymous}.\footnote{We make no distinction between polysemy, homonymy, and other sources of lexical ambiguity a word may exhibit.}
\newcite{dautriche2015weaving} estimates that about 4\% of word forms are homophones: ``such variation is the rule rather than the exception'' %
\citep{cruse1986lexical}. \looseness=-1

Lexical ambiguity is, in general, a fuzzy concept. 
Not only can it be unclear what it means for two senses to be distinct, but different linguistic annotators will also have different opinions on what constitutes a word sense versus a productive use of metaphor. Often the 2\textsuperscript{nd} or 3\textsuperscript{rd} definitions of a word in a dictionary blur this line \citep{LakoffJohnson80}---in \wordnet{} \citep{wordnet}, for instance, the third sense of \word{attack} (intense adverse criticism, e.g.\ ``the government has come under attack'') could be viewed as a metaphorical usage of the first (a military offensive against an enemy, e.g.\ ``the attack began at dawn''), projected from one domain to another. Indeed, this fuzziness has led some researchers to prefer unsupervised word sense induction methods, as they obviate the potentially problematic annotation altogether 
\citep[e.g.][]{panchenko}.
Such unsupervised methods are not without problems, though, with one example being their overreliance on topical words \cite{amrami2019towards}.
These difficulties motivate us to opt for using two distinct representation of a word's lexical ambiguity: one hand-annotated and discrete, the other unsupervised and continuous.

\vspace{-5pt}
\subsection{Accounts of Lexical Ambiguity}
\vspace{-2pt}

When investigating the relationship between ambiguity and word frequency, \citeauthor{zipf1949human} argued that ambiguity results as a trade-off
from opposing forces between speaker and listener, together optimising the communication channel via a principle of least effort: the listener wants to easily disambiguate, the speaker wants to choose words which required little effort to utter, and to avoid excessively searching their lexicon. 
\looseness=-1

Building on \citeposs{zipf1949human} theories, \newcite{piantadosi2012communicative} posit that, when viewed information-theoretically,
ambiguity is in fact a \emph{requirement} for a communication system to be efficient. 
Focusing on economy of expression, \citeauthor{piantadosi2012communicative}\ suggest that %
lexical ambiguity serves a purpose when the context allows for disambiguation---it allows the re-use of simpler word forms.%
\footnote{Recent work, though, has shed some doubt in the interpretation behind these results, showing they might arise solely due to a language's phonotactics distribution \cite{trott2020human,caplan2019miller}.}
They support their hypothesis by demonstrating a correlation between the number of senses for a word listed in \wordnet{} \citep{wordnet} and a number of measures of speaker effort---phonotactic well-formedness, word length and the word's log unigram probability (based on a maximum-likelihood estimate from a large corpus).
\looseness=-1

More recently, \newcite{dautriche2018learning} showed that languages' homophones are more 
likely to appear across distinct syntactic and semantic categories, and will therefore be naturally easier to disambiguate. %
In this work, we show that speakers compensate for lexical ambiguity by making contexts themselves more informative in its presence.

We note an important detail in one of \citeauthor{piantadosi2012communicative}'s experiments. 
In their work, they employ unigram surprisal (i.e. $-\log \unigram(\cdot)$,
where $\unigram(\cdot)$ is the unigram distribution) as a proxy for ease of production,
correlating this with polysemy. 
They justify this approximation based on the fact that more frequent words are, in general, processed more quickly \citep{reder1974semantic}. 
However, this measure has a confounder with our hypothesis: a word's frequency correlates with its contextual uncertainty.
We believe our proposed measure to be more directly connected with lexical ambiguity.\looseness=-1

\vspace{-1pt}
\section{Ambiguity and Uncertainty} 
We formulate both lexical ambiguity and contextual uncertainty information-theoretically. Let $\calM$ be a space of all lexical meaning representations, $\calW$ be the space of all words and $\calC$ be the space of all contexts. We denote the $\calM$-, $\calW$-, and $\calC$-valued random variables as $M$, $W$ and $C$, respectively, and name elements of those sets $m$, $w$ and $c$. %
We take $\calM$ to be an either discrete or continuous meaning space,
$\calW$ to be the set of words in a language 
(excluding the beginning-of- and end-of-sequence symbols, $\bos$ and $\eos$) and
\begin{align} \label{eq:def_context}
    \calC = \{\langle \bos \circ \vp, \vs \circ \eos \rangle \mid \vp \circ w \circ \vs \in \calW^* \}
\end{align}
where $\circ$ denotes string concatenation, and $\vp$ and $\vs$ are the prefix and suffix context strings respectively. 
This set contains every possible context that could surround a word, padded with beginning-of-sequence and end-of-sequence symbols. We additionally define $\widetilde{\vp} = \bos\circ \vp$ and $\widetilde{\vs} = \vs \circ \eos$.

\subsection{Lexical Ambiguity}

We start with a formalisation of lexical ambiguity. Specifically, we formalise the lexical ambiguity of \emph{an entire language} as%
\begin{align}
    \ent(M &\mid W) =  \\
    &-\sum_{w \in \calW} p(w) \int p(m \mid w) \log_2 p(m\mid w)\, \dif m \nonumber
\end{align}
Interpreting entropy as uncertainty, this definition implies that the harder it is to predict the meaning of a word from its form alone, the more lexically ambiguous that word must be. 

We will generally be interested in the half-pointwise entropy, rather than the entropy itself.
In the case of lexical ambiguity, we consider the following half-pointwise entropy%
\begin{align} 
    \ent(M \mid\,&W=w) = \label{eq:polysemy} \\
    &- \int p(m \mid w) \log_2 p(m\mid w)\, \dif m \nonumber
\end{align}
This half-pointwise entropy tells us how difficult it is to predict the meaning when you know the specific word \emph{without} considering its context. We will not generally have access to 
the true distribution $p(m \mid w)$, so we will need
to approximate this entropy. This is discussed in \cref{sec:approximation_entorpy}.
    A unique feature of this operationalisation of lexical ambiguity is that it is language independent.\footnote{We acknowledge the abuse of this bigram in the NLP literature \cite{bender2009linguistically}, and use it in the following specific sense: the operationalisation may be applied to \emph{any} language independent of its typological profile.} However, the quality of a possible approximation will vary from language to language, depending on the models and the data available in that %
language.

A final note %
is that mutual information between $M$ and $W$ \emph{as a function of} $w$ is equivalent, up to an additive constant, to the conditional entropy%
\begin{equation}\label{eq:word_meaning_info}
    \MI(M; W=w) = \ent(M) - \ent(M \mid W=w)
\end{equation}
where $\ent(M)$ is constant with respect to $w$.\footnote{\color{purple}
\textbf{Correction:} The formula given in \Cref{eq:word_meaning_info} is not correct. 
See Proposition 2 in \citet{du2024context}.
Specifically, $\ent(M)$ should be replaced with the following pointwise cross-entropy $\ent_w(M) = - \int p(m \mid w) \log_2 p(m)\, \dif m$.
We do not expect this error to affect our empirical findings.
}\footnote{The mutual information between a discrete and a continuous random variable does not always decompose as the difference between two entropies.
However, it does if we make the good-pair assumption \citep{beknazaryan2018mutual}, which we (implicitly) do throughout this paper.}

This equation asserts something rather trivial: that lexical ambiguity is inversely correlated with how informative a word is about its meaning.

\subsection{Contextual Uncertainty}\label{sec:context-pred}
The predictability of a word in context is also naturally operationalised information-theoretically. We take the contextual uncertainty, once again defined for \emph{an entire language}, as%
\begin{align}
\ent(W &\mid C) = \\
&-\sum_{w \in \calW} p(w) \sum_{\vc \in \calC} p(\vc \mid w) \log_2 p(w \mid \vc) \nonumber
\end{align}
Again, we are mostly interested in the half-pointwise entropy, which tells us how predictable a given word is, averaged over all contexts:
\begin{align} \label{eq:surprisal_context}
\ent(W=w &\mid C) = \\
& - \sum_{\vc \in \calC} p(\vc \mid w) \log_2 p(w \mid \vc) \nonumber
\end{align}%

We take this as our operationalisation of contextual uncertainty.
We note that this definition is different to typical uses of surprisal in computational psycholinguistics \citep{hale, levy2008expectation,seyfarth2014word,piantadosi2011word,pimentel-etal-2020-phonotactic}.
Most work in this vein attempts to maintain cognitive plausibility, usually calculating surprisal based on only the unidirectional left piece of the context, as $-\log p(w \mid \vc_{\leftarrow})$.

Although surprisal is the operationalisation we are interested in here, 
we note that a word may have low surprisal if it is frequent \emph{across many} contexts and not just in a specific one under consideration. 
Sticking with our notion of half-pointwiseness, we define contextual informativeness as%
\begin{equation}
\begin{aligned} \label{eq:mi_context}
    \MI(W&=w; \,C)= \\
   & \ent(W=w) - \ent(W =w \mid C)  
\end{aligned}
\end{equation}
where we define a word's pointwise entropy (also known as surprisal) as
\begin{equation}
\ent(W=w) = - \log_2 p(w)
\end{equation}
The mutual information between a word and its context was studied before by 
\newcite{bicknell2011readers}, \newcite{futrell-levy-2017-noisy} and \newcite{futrell2020lossy}---although only using the unidirectional left piece of the context. \Cref{eq:mi_context} again asserts something trivial: low contextual uncertainty implies in an informative context. This informativeness itself is upper-bounded by the word's absolute negative log-probabiliy (i.e. the unigram surprisal).

\section{Hypothesis: Why Should Ambiguity Correlate with Uncertainty?} \label{sec:robustness}

As discussed in \cref{sec:introduction}, we expect the linguistic signal to be on average somewhat overdetermined or redundant---such redundancy leads to \defn{robustness} in noisy situations, when part of the signal may be lost during its implementation.
A natural measure of robustness is the three-way mutual information between the context of a word, the word itself, and meaning---$\MI(M ; C ; W)$---which represents how much information about the meaning is redundantly encoded in both the context and the word.
The half-pointwise tripartite mutual information can be decomposed as%
\begin{subequations}
\begin{align}
    \MI(M& ; C ; W=w) \\
    &= \MI(M ; W=w) - \MI(M ; W = w \mid C) \nonumber \\
    &= \MI(M; W=w) - \ent(W=w \mid C)  \\
    & \quad\quad  + \cancel{\ent(W=w \mid M, C)} \\
    &\approx \underbrace{\MI(M; W=w)}_{(1)} - \underbrace{\ent(W=w \mid C)}_{(2)} 
\end{align}
\end{subequations}
In this equation, we assume there are no true synonyms under a specific context---i.e.\ given a meaning and a context there is no uncertainty about the word choice: $\ent(W=w \mid M, C) \approx 0$. Term 1 is the information a word shares with its meaning (which is inversely correlated with lexical ambiguity; see \cref{eq:word_meaning_info}) and term 2 is the predictability of a word in context or the contextual uncertainty (which is itself inversely correlated with contextual informativeness; see \cref{eq:mi_context}).

For a language to be efficient, it may reuse its optimal word forms (as defined by their utterance effort), increasing lexical ambiguity \cite{piantadosi2012communicative} and reducing the amount of information a word contains about its meaning (term 1).
This reduces redundancy though, %
increasing the chance of miscommunication in the presence of noise.
Speakers can compensate for this by making contexts more informative for these words (term 2 smaller).
A negative correlation between contextual uncertainty and lexical ambiguity then arises from the trade-off between clarity and economy.

\section{Computation and Approximation}\label{sec:approximation}
Our information-theoretic operationalisation requires approximation. First, we do not know the true distributions over words, their meanings and their contexts. Second, even if we did, \cref{eq:polysemy} and \cref{eq:surprisal_context} would likely be hard to compute.

\subsection{Lexical Ambiguity} \label{sec:approximation_entorpy}

In this section, we provide two approximations for lexical ambiguity. One assumes discrete word senses and requires data annotation ($\wordnet{}$), while the other considers continuous meaning spaces ($\BERT{}$) and allows us to extend our analysis to languages with fewer of these resources.

\vspace{2.5pt}
\paragraph{Discrete senses}
$\wordnet{}$ \citep{wordnet} is a valuable resource available in high-resource languages, which provides a list of synsets for word types.
By taking these synsets to be the possible meanings of a word, and assuming a uniform distribution over them, we approximate the entropy as
\begin{equation}\label{eq:polysemy_wordnet}
    \ent(M \mid W=w) \approx \log_2(\#\textit{senses}[w])
\end{equation}

\vspace{2pt}
\paragraph{Continuous meaning space} 
We now describe how to approximate
ambiguity using $\BERT{}$ \cite{devlin2019bert}.\footnote{We used the implementation of Multilingual $\BERT{}$ made available by \newcite{wolf2019hugging}.}
Let $w \in \calW$ be a word and let $\vc = \langle \widetilde{\vp}, \widetilde{\vs} \rangle \in \calC$ be a padded context. We assume that a word's contextual embedding in $\BERT{}$ (i.e.\ its final hidden state) is a good approximation for its meaning in a given sentence.\footnote{Since $\BERT{}$ returns embeddings for WordPiece units \citep{wu2016google} rather than words, we average them per word to get embeddings at the word-level. We acknowledge that this is a na\"{i}ve method of compositionality; improving the method would likely strengthen our results.} We define the hidden state of a word $w$ in a context $\vc$ as
\begin{equation} \label{eq:hiddenstate}
    \vh_{\langle w, \vc \rangle} = \BERT(\widetilde{\vp} \circ w \circ \widetilde{\vs})
\end{equation}
and we approximate the true distribution over
words, meanings and contexts by %
\begin{equation}\label{eq:approx-dist}
    p(w, m, \vc) \approx \delta(m \mid w, \vc)\,p(w, \vc)
\end{equation}
where we define %
$\delta(m \mid w, \vc)$ to place probability 1 on the point $m =  \vh_{\langle w, \vc \rangle}$ and 0 on every other point.
In other words, we assume the meaning is a deterministic function of a word--context pair, and that it is approximated by $\BERT{}$'s hidden state.

This alone is not enough to estimate \cref{eq:polysemy}, though, since we still do not have access to the true distribution $p(w, \vc)$.
Furthermore, estimating the marginal distribution $p(m|w)$ directly is infeasible, given the sparsity of the meaning space. 
Instead, we approximate an upper bound of the entropy directly---exploiting the fact that a Gaussian distribution $\normal(\mu, \Sigma)$ will have an entropy that is greater than or equal to any other distribution with the same finite and known (co)variance \cite[Chapter 8]{cover2012elements}:\footnote{We note that, unlike its discrete counterpart, differential entropy values can be negative.}%
\begin{align}\label{eq:entropy-bound}
    \ent(M &\,\mid W=w) \\
    &\le \ent(\normal(\vmu_w, \Sigma_w)) = \frac{1}{2}\log_2 \det\left(2\pi e \Sigma_w \right) \nonumber
\end{align}
We estimate this covariance based on a corpus of $N$ word--context pairs $\{\langle w_i, \vc_i \rangle\}_{i=1}^{N}$, which we assume to be sampled according to the true distribution $p$ (our corpora comes from Wikipedia dumps and is described in \cref{sec:data}).\footnote{We explain how to approximate the covariance matrix $\Sigma_w$ per word type in  \cref{sec:bert_gaussian_aprox}.}

The tightness of this upper bound on the entropy depends on both the accuracy of the covariance matrix estimation and the nature of the true distribution $p(m \mid w)$. 
If $p(m \mid w)$ is concentrated in a small region of the meaning space (corresponding to a word with nuanced implementations of the same sense), %
the bound in \cref{eq:entropy-bound} could be relatively tight. 
In contrast, a word with several unrelated homophones would correspond to a highly structured $p(m \mid w)$ (e.g.\ with multiple modes in far distant regions of the space) for which this normal approximation would result in a very loose upper bound. %
\looseness=-1

\subsection{Contextual Uncertainty}

How uncertain the context is about a specific word is formalised in the half-pointwise entropy presented in \cref{eq:surprisal_context}.
We may get an upper bound on this entropy from its cross-entropy:
\begin{align} \label{eq:surprisal_upper}
    \ent(W=w \mid C) &\le \ent_{q_{\boldsymbol\theta}}(W=w \mid C) \\
    &=  - \sum_{\vc \in \calC} p(\vc \mid w) \log q_{\boldsymbol\theta}(w \mid \vc) \nonumber
\end{align}
where $q_{\boldsymbol\theta}$ is a cloze language model that we train to approximate $p$ (as we explain later in this section).
This equation, though, still requires an infinite sum over $\calC$. 
We avoid that by using an empirical estimate of the cross-entropy:
\begin{align}
    \ent_{q_{\boldsymbol\theta}}(W=w \mid C) \approx - \sum_{i = 1}^{N_w} \log q_{\boldsymbol\theta}(w_i \mid \vc_i)
\end{align}
where $N_w$ is the number of samples we have for a specific word type $w$.

To choose an appropriate distribution
$q_{\boldsymbol\theta} (w \mid \vc)$, we train a model on a masked language modelling task. Defining $\mask$ as a special type in vocabulary $V$, we take a masked hidden state as
\begin{equation} \label{eq:masked-hidden-state}
    \vh_{\vc} = \BERT(\widetilde{\vp} \circ \mask \circ \widetilde{\vs})
\end{equation}
We then use this masked hidden state to estimate the distribution
\begin{align} \label{eq:langmod}
     q_{\boldsymbol\theta} (w\mid \vc) &= \softmax (W^{(2)} \sigma (W^{(1)} \vh_{\vc}))_w
\end{align}
where $W^{(\cdot)}$ are linear transformations, and bias terms are omitted for brevity. We fix $\BERT{}$'s parameters and train this model with Adam \cite{kingma2014adam}, using its default learning rate in PyTorch \citep{pytorch2019}. 
We use a ReLU as our non-linear function $\sigma$ and $200$ as our hidden size, training for only one epoch.
By minimising cross-entropy loss we achieve an estimate for $p$.

We do not use $\BERT$ directly as our model $q_{\boldsymbol\theta}$ because its multilingual version was trained on multiple languages, and, thus, was not optimised on each individually.
We found this resulted in poor approximations on the lowest-resource languages.
Furthermore, we note that $\BERT$ gives probability estimates for word pieces (as opposed to the words themselves), and combining these piece-level probabilities to word-level ones is non-trivial.
Indeed, doing so would require running $\BERT$ several times per word, increasing the already high computational requirements of this study. 
To compute the probability of a word composed of two word pieces, for example, we would need to run the model with two masks, i.e. $\BERT(\widetilde{\vp} \circ \mask \circ \mask \circ \widetilde{\vs})$, and combine the pieces' probabilities.
To correctly estimate the probability distribution over the entire vocabulary (i.e.\ $q_{\boldsymbol\theta} (w \mid \vc)$), we would need to replace each position with an arbitrary number of $\mask$s and normalise these probability values.
\looseness=-1

\vspace{-2pt}
\section{Data}\label{sec:data}

We used Wikipedia
as the main data source for all our experiments.
Multilingual $\BERT{}$\footnote{Information about multilingual $\BERT$ can be found in: \url{https://github.com/google-research/bert/blob/master/multilingual.md}}
was trained on the 104 languages with the largest Wikipedias\footnote{List of Wikipedias can be found in \url{https://meta.wikimedia.org/wiki/List_of_Wikipedias}}---of these, we subsampled a diverse set of 18 for our experiments: 
Afrikaans, Arabic, Bengali, English, Estonian, Finnish, Hebrew, Indonesian, Icelandic, Kannada, Malayalam, Marathi, Persian, Portuguese, Tagalog, Turkish, Tatar, and Yoruba.
\looseness=-1

For each of these languages, we first downloaded their entire Wikipedia, which we sentencised and tokenizsd using language-specific models in spaCy \cite{spacy}---our definition of a word here is, thus, a token as given by the spaCy tokeniser.
We then subsampled 1 million random sentences per language for our analysis and another $100{,}000$ random sentences to train the model $q_{{\boldsymbol\theta}}$.
We run multilingual $\BERT{}$ on the 1 million analysis sentences to acquire both $\vh_{\langle w, \vc \rangle}$ and $\vh_{\vc}$ (\cref{eq:hiddenstate} and \cref{eq:masked-hidden-state}) for each word in these corpora---discarding any word for which we do not have at least 100 contexts in which the word occurs. 
For the purpose of our analysis, we also discarded any word containing characters not in the individual scripts of the analysed language.
The final number of word types used in our analysis can be found in \cref{tab:polysemy,tab:correlations-full}.
\looseness=-1

\begin{figure}
    \centering
    \includegraphics[width=\columnwidth]{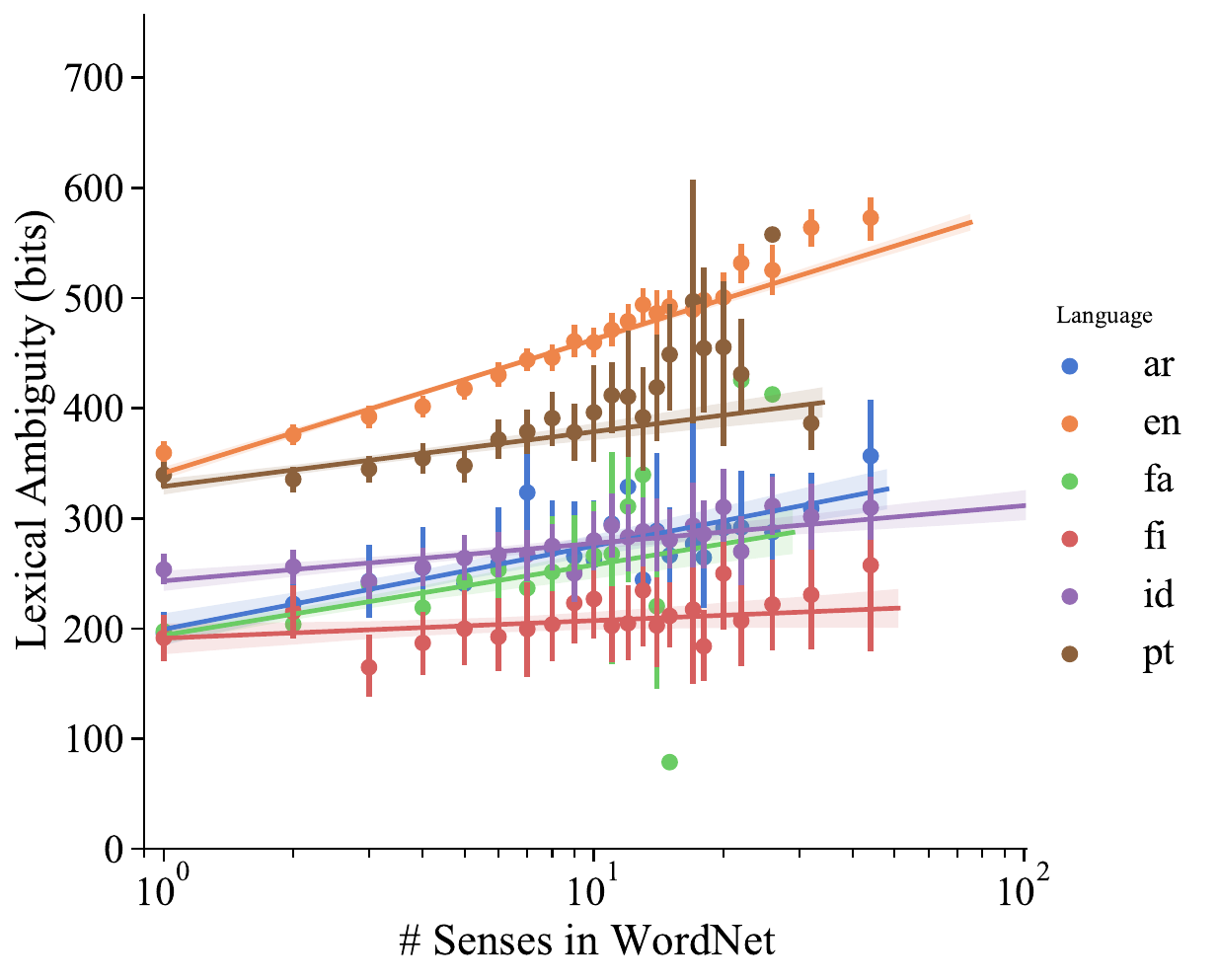}
    \caption{Correlating our $\BERT$-based estimate of lexical ambiguity with the number of senses in \wordnet{}}%
    \label{fig:en_polysemy_corr}
\end{figure}

\section{Discussion: \wordnet{} vs.\ $\bm{\BERT{}}$-based approximations}

The novel continuous ($\BERT$-based) approximation of lexical ambiguity 
has two important virtues over the alternative \wordnet{}-based measure. 
On the practical side, it can be readily computed for many languages. 
Since we are using multilingual $\BERT{}$ for our continuous approximation, as discussed in \cref{sec:approximation}, this quantity is easily obtainable for the 104 languages on which it was trained. 
Second, on more theoretical grounds, the continuous representation of the space of meanings might better capture the gradient that goes from subtle but distinct senses of the same word to completely unrelated homophones \cite[p. 51]{cruse1986lexical}.
Alternatively, the \wordnet{}-based measure of lexical ambiguity is supported by expert human annotation and extensive research on its linguistic and psycholinguistic correlates, e.g.\ \newcite{sigman2002global} and \newcite{budanitsky2006evaluating}.
These differences notwithstanding, we expect both measures to correlate to a certain degree. %
To evaluate this, we run an experiment comparing both estimates in six languages from four different families for which \wordnet{} is available: Arabic, English, Finnish, Indonesian, Persian, and Portuguese.\looseness=-1

\begin{table}
    \centering
    \begin{tabular}{l c c c}
    \toprule
        Language & \# Types & Pearson & Spearman \\
    \midrule
Arabic & 836 & 0.25$^{**}$  & 0.30$^{**}$ \\
English & 6995 & 0.40$^{**}$  & 0.40$^{**}$ \\
Finnish & 1247 & 0.06$^*$~~  & 0.07$^*$~~ \\
Indonesian & 3308 & 0.12$^{**}$  & 0.13$^{**}$ \\
Persian & 2648 & 0.14$^{**}$  & 0.13$^{**}$ \\
Portuguese & 3285 & 0.13$^{**}$  & 0.13$^{**}$ \\
    \bottomrule 
\multicolumn{3}{l}{$^{**}$ $p<0.01$ $^{*}$ $p<0.1$}
    \end{tabular}
    \caption{Correlations between a word's lexical ambiguity as estimated with $\BERT{}$ or \wordnet{}.
    }
    \label{tab:polysemy}
\end{table}

\cref{fig:en_polysemy_corr} and \cref{tab:polysemy} show that indeed both measures are positively correlated, although the association may be modest in some languages.
The Pearson correlation between our estimates is $\rho=0.40$ for English, but only $\rho=0.06$ for Finnish---other languages lie in the range between the two.\footnote{For all tests of significance in this paper, we apply \citeauthor{benjamini1995controlling}'s correction (\citeyear{benjamini1995controlling}).}
This correlation seems to increase with the quality of the $\BERT$ model for the language under consideration---English has the largest Wikipedia, so multilingual $\BERT{}$ should naturally be better modelling it, while Finnish has the smallest Wikipedia among these six languages.
A complementary explanation is that \wordnet{} itself might be better for English than other languages---while English's \wordnet{} contains synsets for $147{,}306$ words, Persian only has them for $17{,}560$.
This suggests that the modest associations found should be taken as pessimistic lower bounds.\looseness=-1

A potential underlying problem in the above study is that the number of senses a word has in \wordnet{} might rely on word frequency (this beyond a true underlying relationship with it)---e.g. annotating senses for frequent words may be easier than for infrequent ones. 
Furthermore, the number of samples a word has in our corpus will affect its sample density in the embedding space and thus its estimated $\BERT$ entropy.
As a second evaluation,
we therefore train a multivariate linear regressor predicting our $\BERT$-based measure not only from the log of the number of senses a word has in \wordnet{}, but also the word's frequency (i.e.\ its number of occurrences in the corpus). This analysis is presented in \cref{tab:polysemy_multivariate}, where we can see that both our estimates of lexical ambiguity still correlate when controlling for frequency. This table also shows that our $\BERT$-based estimate still correlates with the word's frequency when controlling for the number of senses the word has in \wordnet{}. Future work could delve further into what this correlation implies, with the potential to improve our proposed annotation-free estimate of lexical ambiguity.
\looseness=-1

\begin{table}
    \centering
    \begin{tabular}{l c c c}
    \toprule
        Language & \# Types & \wordnet{} & Frequency \\
    \midrule
Arabic & 836 & 0.28$^{**}$ & 0.30$^{**}$ \\    
English & 6995 & 0.38$^{**}$ & 0.21$^{**}$ \\   
Finnish & 1247 & 0.07$^*$~~ & 0.35$^{**}$ \\    
Indonesian & 3308 & 0.09$^{**}$ & 0.37$^{**}$ \\             
Persian & 2648 & 0.13$^{**}$ & 0.14$^{**}$ \\ 
Portuguese & 3285 & 0.13$^{**}$ & 0.29$^{**}$ \\
    \bottomrule 
\multicolumn{3}{l}{$^{**}$ $p<0.01$ $^{*}$ $p<0.1$}
    \end{tabular}
    \caption{
    Parameters (and their significance) of a multivariate linear regression predicting our $\BERT{}$-based measure of ambiguity from both our \wordnet{} estimate and the word's frequency. All analysed variables were normalised to have zero mean and unit variance.
    }
\vspace{-15pt}
    \label{tab:polysemy_multivariate}
\end{table}

\section{Lexical Ambiguity Correlates With Contextual Uncertainty} \label{sec:mi_results}
\begin{figure*}
    \centering
    \begin{subfigure}[t]{0.33\textwidth}
        \centering
        \includegraphics[width=\columnwidth]{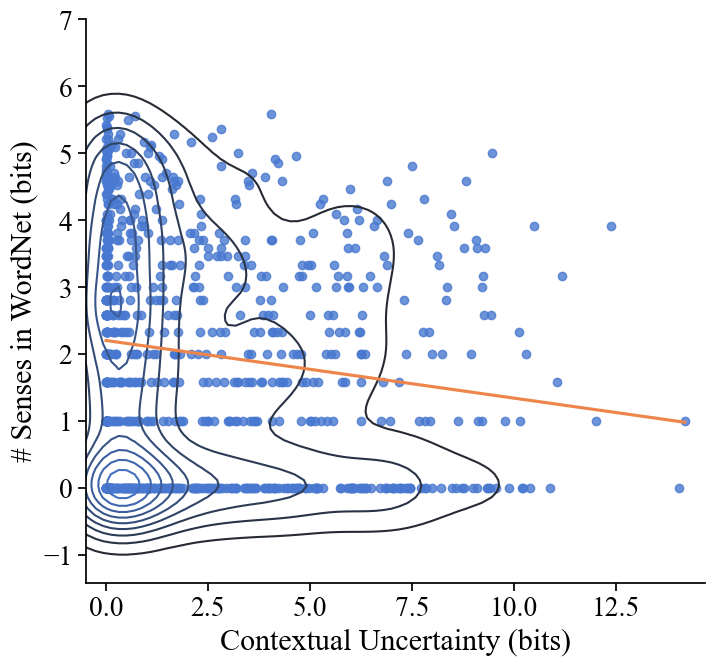}
    \end{subfigure}%
    ~ 
    \begin{subfigure}[t]{0.33\textwidth}
        \centering
        \includegraphics[width=\columnwidth]{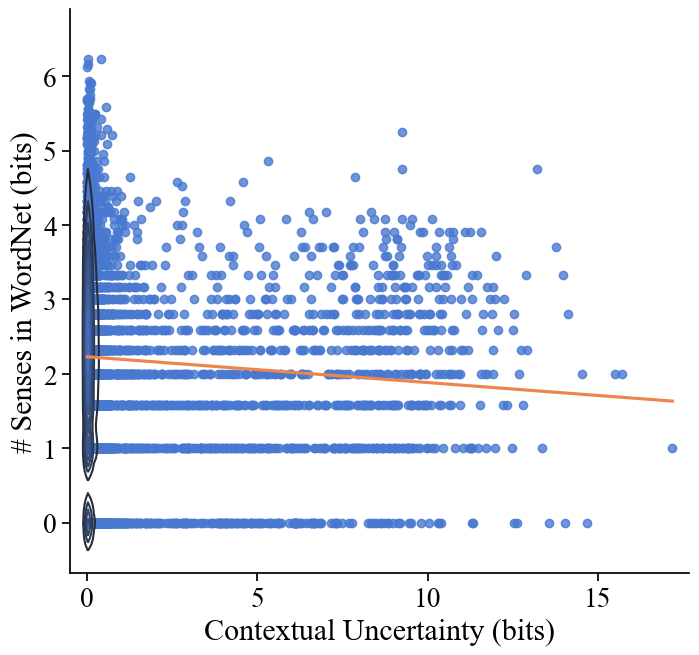}
    \end{subfigure}%
    ~ 
    \begin{subfigure}[t]{0.33\textwidth}
        \centering
        \includegraphics[width=\columnwidth]{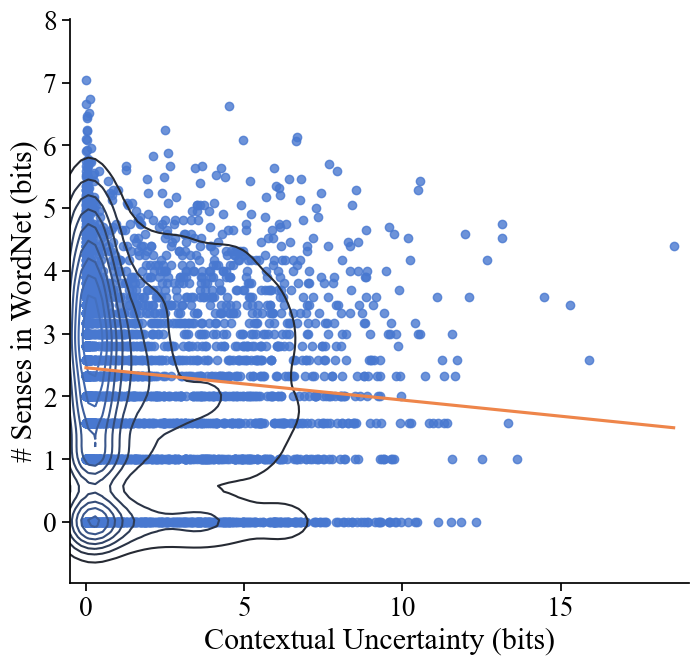}
    \end{subfigure}
    
    \begin{subfigure}[t]{0.24\textwidth}
        \centering
        \includegraphics[width=\columnwidth]{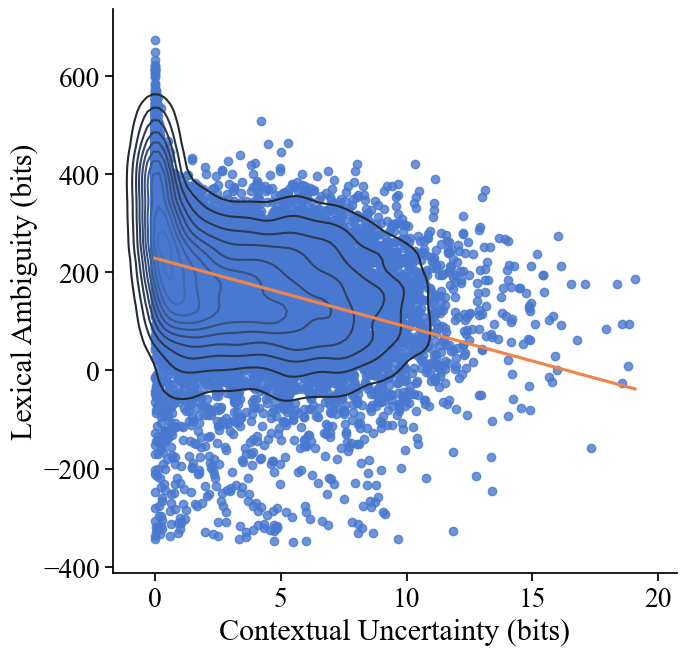}
    \end{subfigure}%
    ~ 
    \begin{subfigure}[t]{0.24\textwidth}
        \centering
        \includegraphics[width=\columnwidth]{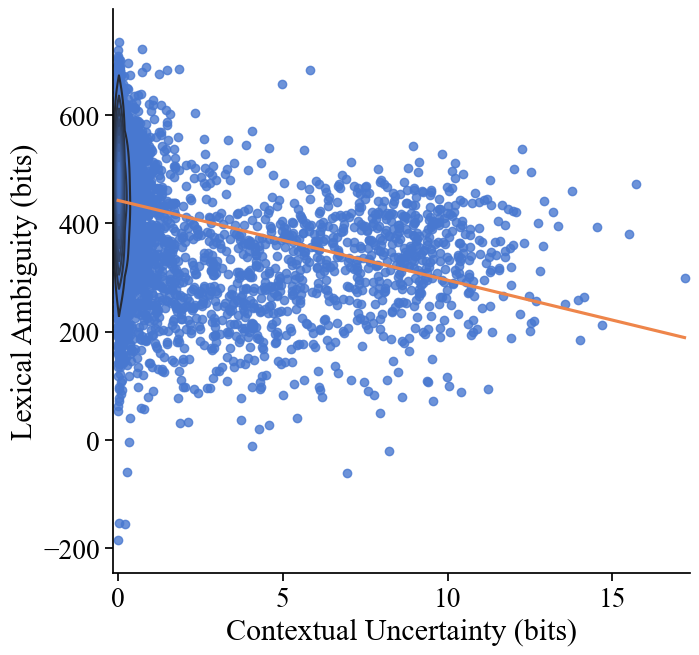}
    \end{subfigure}%
    ~ 
    \begin{subfigure}[t]{0.24\textwidth}
        \centering
        \includegraphics[width=\columnwidth]{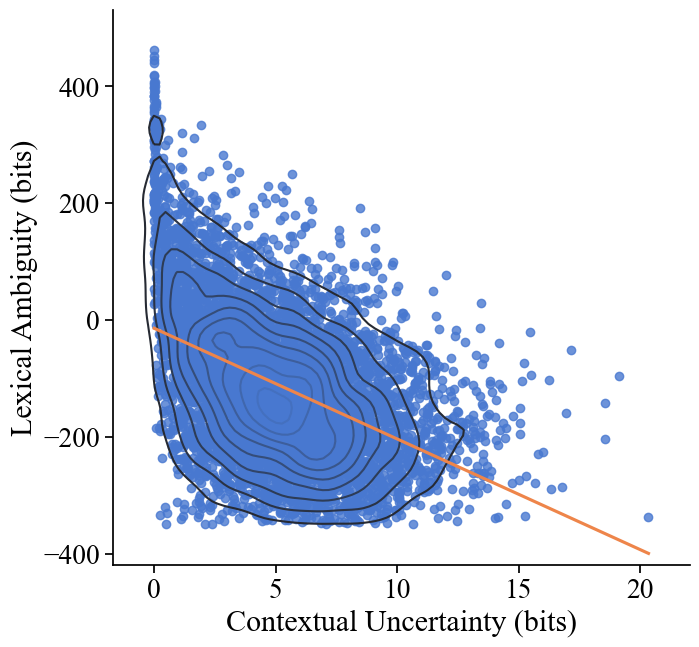}
    \end{subfigure}%
    ~ 
    \begin{subfigure}[t]{0.24\textwidth}
        \centering
        \includegraphics[width=\columnwidth]{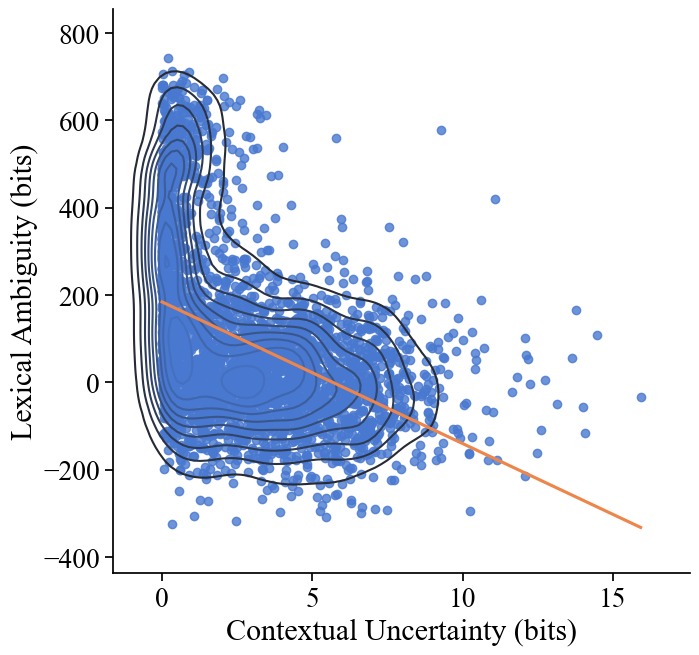}
    \end{subfigure}
    \caption{Contextual uncertainty versus lexical ambiguity %
    in a selection of languages. Each plot contains the scatter points (representing each word type), a robust linear regression and kernel density estimate regions.
    (From left to right; Top) \wordnet{}: Arabic, English, Indonesian;
    (Bottom) $\BERT$: Arabic, English, Malayalam, Tagalog.
    }
    \vspace{-15pt}
    \label{fig:suprisals_corr}
\end{figure*}

We now test whether lexical ambiguity negatively correlates with contextual uncertainty, the main hypothesis of our paper. We first evaluate this on a set of six high-resource languages, using our \wordnet{} estimate for the lexical ambiguity of a word. The top half of \cref{tab:correlations-full} shows the results: for five of the six languages, there is a negative correlation between the number of senses of a word and contextual uncertainty ($p<0.01$). The top half of \cref{fig:suprisals_corr} further presents these results. In these Figures we see that, especially for highly ambiguous words, contextual uncertainty tends to be very small.
This supports our hypothesis, but only on a restricted set of languages for which \wordnet{} is available. 
\looseness=-1

\begin{table}[t]
    \centering
    \resizebox{\columnwidth}{!}{%
    \begin{tabular}{l r c c}
    \toprule
Language & \# Types & Pearson & Spearman \\
    \midrule
\multicolumn{4}{l}{\emph{Lexical ambiguity as \wordnet{}}} \\ \cmidrule(l{.5em}){1-2}
Arabic (ar) & 836 & -0.14$^{**}$  & -0.15$^{**}$ \\
English (en) & 6995 & -0.07$^{**}$  & -0.11$^{**}$ \\
Finnish (fi) & 1247 & \phantom{-}0.01~~~  & -0.00~~~ \\
Indonesian (id) & 3308 & -0.09$^{**}$  & -0.14$^{**}$ \\
Persian (fa) & 2648 & -0.11$^{**}$  & -0.12$^{**}$ \\
Portuguese (pt) & 3285 & -0.10$^{**}$  & -0.11$^{**}$ \\
    \midrule
\multicolumn{4}{l}{\emph{Lexical ambiguity as $\BERT$}} \\ \cmidrule(l{.5em}){1-2}
Afrikaans (af) & 4505 & -0.41$^{**}$  & -0.52$^{**}$ \\
Arabic (ar) & 10181 & -0.33$^{**}$  & -0.41$^{**}$ \\
Bengali (bn) & 8128 & -0.43$^{**}$  & -0.44$^{**}$ \\
English (en) & 7097 & -0.33$^{**}$  & -0.35$^{**}$ \\
Estonian (et) & 4482 & -0.40$^{**}$  & -0.44$^{**}$ \\
Finnish (fi) & 3928 & -0.38$^{**}$  & -0.45$^{**}$ \\
Hebrew (he) & 13819 & -0.34$^{**}$  & -0.37$^{**}$ \\
Indonesian (id) & 4524 & -0.45$^{**}$  & -0.57$^{**}$ \\
Icelandic (is) & 3578 & -0.44$^{**}$  & -0.46$^{**}$ \\
Kannada (kn) & 9695 & -0.42$^{**}$  & -0.41$^{**}$ \\
Malayalam (ml) & 6203 & -0.47$^{**}$  & -0.46$^{**}$ \\
Marathi (mr) & 5821 & -0.39$^{**}$  & -0.40$^{**}$ \\
Persian (fa) & 6788 & -0.39$^{**}$  & -0.49$^{**}$ \\
Portuguese (pt) & 5685 & -0.31$^{**}$  & -0.45$^{**}$ \\
Tagalog (tl) & 3332 & -0.45$^{**}$  & -0.50$^{**}$ \\
Turkish (tr) & 4386 & -0.40$^{**}$  & -0.46$^{**}$ \\
Tatar (tt) & 2997 & -0.34$^{**}$  & -0.39$^{**}$ \\
Yoruba (yo) & 417 & -0.55$^{**}$  & -0.64$^{**}$ \\
    \bottomrule 
 \multicolumn{3}{l}{$^{**}$ $p<0.01$}
    \end{tabular}
    }
    \caption{Correlation between lexical ambiguity and contextual uncertainty. 
    }
    \label{tab:correlations-full}
    \vspace{-15pt}
\end{table}

With that in mind, we now consider a larger and more diverse set of 18 languages, analysed using our $\BERT$-based estimator of lexical ambiguity.
Figures~\ref{fig:mi_full} and \ref{fig:suprisals_corr} show the relationship between contextual uncertainty and lexical ambiguity---in all 18 analysed languages, we find negative correlations, further supporting our hypothesis. 
These correlations are presented in the bottom half of \cref{tab:correlations-full}, and range from Pearson $\rho=-0.31$ in Portuguese to $\rho=-0.55$ in Yoruba ($p<0.01$).

Comparing the top and bottom half of \cref{tab:correlations-full}, we see that the correlations are larger when using our $\BERT$ estimate rather than the \wordnet{} one. We believe this may result from one or all of the following: (i) there is a confounding effect caused by the use of the same model ($\BERT{}$) to estimate both ambiguity and surprisal, (ii) the assumption that the senses in \wordnet{} are uniformly distributed may be simplistic, and (iii) our $\BERT$-based ambiguity estimate may capture a more subtle sense of ambiguity
than \wordnet, which may result in a stronger correlation with contextual uncertainty.\footnote{\newcite[p.\ 51]{cruse1986lexical} argues there are two ways in which context affects a word's semantics---selection between units of distinct senses, or contextual modification of a single sense.}
Nonetheless, even if there is a confounding effect in this second batch of experiments (using $\BERT$ to estimate lexical ambiguity), the first batch (with \wordnet{}) has no such confounding factor---providing strong support for our main hypothesis.
\looseness=-1

A quick visual inspection
of \cref{fig:suprisals_corr} indicates this data might be heteroscedastic---it might have unequal variance across distinct ambiguity levels. 
To investigate this, we run \citeposs{white1980heteroskedasticity} test on the uncertainty--ambiguity pairs. This verifies the intuition that this distribution is heteroscedastic for both our \wordnet{} and $\BERT$ measures ($p<0.01$). Future work should investigate the impact of this heteroscedasticity in lexical ambiguity.
\looseness=-1

\paragraph{Limitations}
This work focuses on proposing new information-theoretic approximations for both lexical ambiguity and bidirectional contextual uncertainty and on positing that these two measures should negatively correlate. In this experiment section, 
we tested the hypothesis on a set of typologically diverse languages. Nonetheless, our experiments are restricted to Wikipedia corpora. This data is naturally limited. For instance, while dialog utterances may rely on extra-linguistic clues, sentences in Wikipedia cannot.
Furthermore, due to its ample audience target, the text in Wikipedia may be over descriptive. Future work should investigate if similar results apply to other corpora.

\section{Conclusion}
In this paper we hypothesised that, were a language economical in its expressions \emph{and} clear, then the contextual uncertainty of a word should negatively correlate with its lexical ambiguity---suggesting speakers compensate for lexical ambiguity by making contexts more informative.
To investigate this, we proposed an information-theoretic operationalisation of lexical ambiguity, together with two methods of approximating it, one using \wordnet{} and one using $\BERT$.
We discuss the relative advantages of each, %
and provide experiments using both. %
With our \wordnet{} approximation, we found significant negative correlations between lexical ambiguity and contextual uncertainty in five out of six high-resource languages analysed, supporting our hypothesis in this restricted setting.
With our $\BERT$ approximation, we then expanded our analysis to a larger set of 18 typologically diverse languages and found significant negative correlations between lexical ambiguity and contextual uncertainty in all of them, further supporting our hypothesis that contextual uncertainty negatively correlates with lexical ambiguity.\looseness=-1

\section*{Acknowledgments}

Dami\'{a}n Blasi acknowledges funding from the framework of the HSE University Basic Research Program and is funded by the Russian Academic Excellence Project `5-100'.\looseness=-1

\bibliography{acl2020}
\bibliographystyle{acl_natbib}

\clearpage
\newpage

\onecolumn
\appendix
\section*{Appendices}

\section{A Gaussian Approximation for Meaning}\label{sec:bert_gaussian_aprox}

Given our samples $\{\langle w_i, \vc_i \rangle\}_{i=1}^N$ of word--context pairs (assumed to be drawn from the true distribution $p$), we get the subset of $N_w$ instances of word type $w$. 
We then use an unbiased estimator of the covariance matrix:
\begin{equation}
    \Sigma_w\approx \frac{1}{N_w-1} \sum\limits_{i = 1}^{N_w} \left(\vh_{ \langle w_i, \vc_i \rangle} - \widetilde{\vmu}_w\right)\left(\vh_{\langle w_i, \vc_i \rangle} - \widetilde{\vmu}_w \right)^{\top}  
\end{equation}
where the sample mean is defined as
\begin{equation}
    \widetilde{\vmu}_w \approx \frac{1}{N_w} \sum\limits_{i = 1}^{N_w} \vh_{\langle w_i, \vc_i \rangle} 
\end{equation}
We note that these approximations become exact as $N_w \rightarrow \infty$ due to the law of large numbers.
Because $\vh_{\langle w, \vc \rangle}$ (i.e.\ $\BERT$'s hidden state) is a $768$-dimensional vector, we might not have enough samples to fully estimate $\Sigma_w$.
Thus, we actually approximate this entropy by using only its variance $\diag(\Sigma_w)$. This is still an upper bound on the true entropy
\begin{equation}
    \ent(\normal(\vmu_w, \Sigma_w)) \le \ent(\normal(\vmu_w, \diag(\Sigma_w)))
\end{equation}
The right side of this equation is, then, used as our actual lexical ambiguity estimate.

\section{ISO 639-1 Codes}

In this Section, we present the set of ISO 639-1 language codes we use throughout this paper---in \cref{tab:iso}.

\begin{table}[h]
    \centering
    \begin{tabular}{ll}
    \toprule
         ISO Code & Language \\
         \midrule
af & Afrikaans \\
ar & Arabic \\
bn & Bengali \\
en & English \\
et & Estonian \\
fi & Finnish \\
he & Hebrew \\
id & Indonesian \\
is & Icelandic \\
kn & Kannada \\
ml & Malayalam \\
mr & Marathi \\
fa & Persian \\
pt & Portuguese \\
tl & Tagalog \\
tr & Turkish \\
tt & Tatar \\
yo & Yoruba \\
         \bottomrule
    \end{tabular}
    \caption{ISO Codes and their languages}
    \label{tab:iso}
\end{table}

\end{document}